\pdfoutput=1
\documentclass[twocolumn]{IEEEtran}
\usepackage[english]{babel}
\usepackage[pdftex]{graphicx}
\usepackage{amsmath}
\usepackage{amsthm}
\usepackage{amsfonts}
\usepackage{amssymb}
\usepackage{cite}
\usepackage{array}
\usepackage{algorithm2e}
\usepackage{url}
\usepackage{xcolor}
\usepackage{xspace}
\usepackage{url}
\usepackage{subfig}
\usepackage{balance}
\theoremstyle{definition}

\newtheorem{proposition}{Proposition}

\def\transp{^\intercal}
\newcommand{\refeq}[1]{(\ref{#1})}

\def\reals{\ensuremath{\mathbb{R}}}

\definecolor{NewTextFG}{rgb}{0.0,0.1,0.8}

\newcommand{\revTwo}[1]{{\color{blue}#1}}%
\def\begRevTwo{\color{blue}}
\def\endRevTwo{\color{black}}

\newcommand{\mnote}[1]{\marginpar{\framebox{\bf#1}}}

\title{Binary Matrix Factorization via Dictionary Learning}
\author{Ignacio Ram\'{\i}rez~\IEEEmembership{Member,~IEEE}\thanks{Departamento de Procesamiento de Se\~{n}ales, Instituto de Ingenier\'{\i}a El\'{e}ctrica, Facultad de Ingenier\'{\i}a, Universidad de la Rep\'{u}blica, Uruguay.} }
\begin{document}
\maketitle
\begin{abstract}
Matrix factorization is a key tool in data analysis; its applications include recommender systems, correlation analysis, signal processing, among others. Binary matrices are a particular case which has received significant attention for over thirty years, especially within the field of data mining. Dictionary learning refers to a family of methods for learning overcomplete basis (also called frames) in order to efficiently encode samples of a given type; this area, now also about twenty years old, was mostly developed within the signal processing field.
In this work we propose two binary matrix factorization methods based on a binary adaptation of the dictionary learning paradigm to binary matrices.
The proposed algorithms focus on speed and scalability; they work with binary factors combined with bit-wise operations and a few auxiliary integer ones. Furthermore, the methods are readily applicable to online binary matrix factorization.
Another important issue in matrix factorization is the choice of rank for the factors; we address this model selection problem with an efficient method based on the Minimum Description Length principle.
Our preliminary results show that the proposed methods are effective at producing interpretable factorizations of various data types of different nature.
\end{abstract}%
\section{introduction}
\label{sec:intro}

We consider the problem of approximating a binary matrix $X \in \{0,1\}^{m{\times}n}$ as the product of two other binary matrices $U \in \{0,1\}^{m{\times}p}$ and $V \in \{0,1\}^{n{\times}p}$ plus a third \emph{residual} matrix $E$,
\begin{equation}
X = UV\transp + E.
\label{eq:mf}
\end{equation}
\begRevTwo
\mnote{1.1}The problem of Binary Matrix Factorization (BMF) arises naturally in many problems, in particular in the so called ``association matrices'' where a $X_{ij}=1$ indicates that some object $i$ belongs to group $j$~\cite{bmf-app-biclustering}, or some entity (e.g., a gene) $i$ is present in some species  $j$~\cite{bmf-app-diversity,bmf-app-microbial}, or are related to some type of disease~\cite{bmf-app-tumor}. The latter examples show the increasing relevance of this type of data in genomics. Other similar, growing applications include recommender systems (client-product preferences, etc.).
\endRevTwo
The BMF problem dates back to at least the 1960s~\cite{bmf-oldest} and has been treated extensively
in the last three decades by various research communities, under quite different names. It was first
studied as a combinatorial problem as a particular case of the classic \emph{set covering} problem
(see~\cite{monson95} and references therein). It then received great attention from the data mining
community.
\begRevTwo\mnote{2.2-3} Long known to be an NP-hard problem~\cite{asso}\endRevTwo, the earlier works in this field developed heuristics such as the \emph{tiling} or \emph{tile matching/searching} methods, where binary matrices are decomposed as Boolean or modulo-2 superpositions of rectangular tiles~\cite{proximus,tiling}; the BMF problem was later formulated as a matrix factorization problem in~\cite{bmf07}, with several works following that line since then.
\begRevTwo
Being an NP-hard problem, the quality of the decompositions $(U,V)$ is commonly \mnote{1.2}measured in terms of their \emph{interpretability}, that is whether the columns of $U$ and $V$ exhibit patterns that are intuitive in some sense, or reflect \emph{a priori} knowledge about the problem. For example, in data mining problems, where the pairs $(U_i,V_i)$ are interpreted as \emph{association rules} (for example, a group of clients -- indicated by non-zeros in $U_i$ -- prefers a certain group of products -- indicated by non-zeros in $V_i$);
the examples in this paper are designed  to show this interpretability in a visual way, by studying the results on visual patterns obtained on sets of images.
\endRevTwo
A thorough survey of BMF methods is beyond the scope of this paper; we refer the reader
to~\cite{bmf-comp} for a more in-depth review. We will however mention some works which are
representative of the diversity of formulations and tools that surround the treatment of this
problem, as well as the shortcomings that are common to the current state of the art and that
motivate the development of the tools that we present in this work.

\subsection{Brief overview of Binary Matrix Factorization}

We begin with the ASSO algorithm proposed in~\cite{asso}. The method begins by constructing the so
called \emph{association matrix} $C$, which is a thresholded version of a particular normalization
of the correlation matrix $X{\transp}X$. It then produces a series of increasing rank approximations
by adding to $U$ a column taken from $C$ and searching for a corresponding binary row $V_k$ whose
outer product with $U_k$ minimizes the number of non-zeros in the current approximation error
$E$. This method can be efficiently implemented with bitwise and integer operations. It is also a
popular and simple method with good performance. However, the complexity of each ASSO step is
$O(kn^2m)$, so that it does not scale well in applications where $n$ (the number of samples) is very
large, something very common in current data science problems, which is the target of our work.

\begRevTwo
\mnote{0.1}
\mnote{2.2-3}
Many works approximate the BMF problem by a relaxed (non-convex) Non-negative Matrix Factorization (NMF) problem where $U$ and $V$ are allowed to take on real values, and then map the resulting (approximate) solution to the binary domain using some predefined rule. Examples of this are~\cite{bmf07,bmf13,bmf-nmf2}. In particular, the work~\cite{bmf13} develops a set of BMF \emph{identifiability} conditions, that is, conditions under which the binary factorization of $X$ is unique (up to permutations). However, as in~\cite{bmf07,bmf13}, their solution is an approximation based on a local minima of the NMF problem, so there are no guarantees that the binarized pair $(U,V)$ obtained coincides with the unique solution even if the identifiability conditions are satisfied. 
Being based on non-linear optimization, the NMF methods are  significantly more computationally demanding than binary methods such as ASSO. 
\endRevTwo

The work~\cite{bmf-mp} stands out as an interesting alternative to BMF which formulates the
decomposition of $X$ as a Bayesian denoising problem with a particular prior on the
unobserved \emph{clean} matrix $\hat{X}$ ($X=\hat{X}+E$) and uses a \emph{Message Passing} algorithm
to find the maximum a posteriori estimation of $\hat{X}$. Message Passing is a mature technique
which in the form presented in this work can be quite computationally demanding. Approximate Message
Passing \cite{amp,gamp} techniques have since been developed which may provide significant
efficiency gains to the technique proposed in~\cite{bmf-mp}, but we are currently unaware of any
development in this direction.

An example of a tile-searching heuristic is the Proximus method~\cite{proximus}, which approximates
the first principal left and right binary components of the binary matrix $X$. The method can be
extended to produce a hierarchical representation of the matrix with further rank-$1$ components,
although these do not coincide with additional factors in a rank-$k$ factorization. Contrary to the
above methods, the focus of Proximus is on speed an scalability, and indeed is much faster and
scales better than any of the above methods. Also, as we will see in the next subsection, finding
the first principal component of a binary matrix is closely related to a crucial step in one of the
main dictionary learning methods. e will describe this method in detail later in this document.

\begRevTwo

\subsection{Dictionary learning}

\emph{Dictionary learning} methods were first introduced in \cite{olshausen97}
and later adopted as a powerful extension of the \emph{transform analysis} concept, ubiquitous in
signal processing since the introduction of Fourier analysis and their related discrete variants
(DFT, DCT). In this setting, the matrix $X \in \reals^{m{\times}n}$ consists of $n$ columns of
dimension $m$, and the decomposition obtained $X=DA+E$ is comprised of a \emph{dictionary} $D$, a
matrix of \emph{linear coefficients} $A$, plus a residual matrix $E$.  Dictionary learning methods
are adaptive: given sufficient data samples, they can be trained to efficiently represent such
samples as a linear combination of very few basis elements or ``atoms'' (see~\cite{dl-review} for a
review).  Despite coming from a very different community, dictionary learning methods can be seen as
matrix factorization methods which are specially tailored to the where $X$ is either extremely
``fat'' ($n \gg m$) or ``tall'' ($n \ll m$). Furthermore, many of these methods can be implemented on-line, that is,
they can process new data samples as they arrive, and adapt the dictionary along the
way~\cite{online-dl}.  Another important aspect of dictionary learning methods is that they are not
restricted to low-rank decompositions; if fact, the solutions can be \emph{overcomplete}, meaning that the
number of columns $p$ in $D$ may be (much) larger than $m$, the dimension of the samples to be
represented.

\mnote{2.2-3}As with all BMF methods, dictionary learning problems are non-convex and their solution cannot be obtained exactly. Nevertheless, similarly to what happens with BMF, their enormous practical success in a wide range of signal processing and machine learning problems, and their ability to produce human-interpretable patterns, has led to their widespread adoption. 
\endRevTwo

\subsection{Model selection}

As with any statistical model, the problem of model selection, (in this case, choosing $p$) is of
paramount importance to BMF. Various works~\cite{bmf-mdl,bmf-tiling-mdl,bmf-sel,panda} have
addressed this particular problem. In particular,~\cite{bmf-mdl} and~\cite{bmf-tiling-mdl} are based
on the Minimum Description Length (MDL) principle~\cite{mdl1,mdl2,mdl3}, which forms the basis of
our model selection strategy as well. As a side note, the work~\cite{bmf-tiling-mdl} represents a
recent example of the tiling approach. This problem has also been addressed in dictionary learning
in~\cite{dl-mdl}.

\subsection{Main contribution}

\begRevTwo

To the best of our knowledge, the works in the existing literature on BMF make no particular
assumptions on the \emph{shape} of the matrices to be decomposed.  In particular, many methods which
are efficient for the $n \approx m$ case do not scale well for $n \gg m$ and vice versa. Also, most
methods deal with the \emph{offline} analysis of readily available matrices, which makes them
unsuitable to many recent data processing tasks involving \emph{online} adaptation of the models.

The main motivation behind this work is the ability of dictionary learning methods to cope with the challenges mentioned above. We propose two dictionary learning-based BMF methods  which are particularly suited to the treatment of extremely fat (or tall) matrices, and which are also suitable to online processing of samples. The first one, named Method of Binary Directions (MOB), is an adaptation of the method proposed in~\cite{online-dl}, itself an adaptation of the Method of Optimum Directions (MOD)~\cite{mod} (hence the name). The second method is based on the idea of sequential rank-one updates of the K-SVD method~\cite{aharon06};  we adopt the Proximus~\cite{proximus} algorithm as a fast approximation to the mentioned operation; we thus bring together tools from two inherently related but otherwise disconnected fields to obtain a novel formulation to the binary matrix factorization. Finally, both methods rely on a third novel algorithm which we call Binary Matching Pursuit (BMP). This is a binary adaptation of the Matching Pursuit method~\cite{mp} for approximating the sparsest solution to a least squares regression problem. \mnote{2.1}This adaptation, although conceptually analogous to MP, is non-trivial, as its efficiency relies on a careful combination of different algebraic operations.

\endRevTwo
Our methods construct binary dictionaries and binary coefficients matrices using efficient bitwise and a few integer operations. This is particularly relevant to the efficiency of our methods as recent processor architectures incorporate the ability to handle large number of bits through SIMD (Single Instruction Multiple Data) instructions. For example, a current off-the-shelf processor can perform a \emph{popcount} instruction (which counts the number of $1$s in a binary array) on a 256-bit register. This allows, for example, to compute the dot product between two binary vectors of dimension $256$ with just two processor instructions. \begRevTwo \mnote{2.4}As we show in Section~\ref{sec:bdl}, our methods are also linear both in $n$ and $m$ and thus scale well for large matrices.\endRevTwo

\begRevTwo The main objectives of this paper are to present our proposed methods, assess their interpretability on different datasets, to analyze their computational properties such as computational complexity and convergence rate, and to see how these properties are affected by the initial conditions. We thus focus our experiments on a small set of easily-interpretable datasets for which the patterns obtained can be easily recognized as  salient features in the data. \endRevTwo

The rest of this document is organized as follows: Section~\ref{sec:background:dictionary-learning} provides the notation and background on the methods on which our methods are based. The proposed methods themselves are described in Section~\ref{sec:bdl}.
Section~\ref{sec:model-selection} describes the proposed MDL-based model selection algorithm for searching the best model order $p$. %
We present and discuss our results in Section~\ref{sec:results}, and provide concluding remarks in Section~\ref{sec:conclusion}.

\section{Background}
\label{sec:background}

\subsection{Notation}
\label{sec:background:notation}
%
% NOTATION
%
\def\indicator{\mathbf{1}}
\def\bool{\mathrm{bool}}
\def\bprod{\circ}
\def\bsum{\lor}
\def\bigand{\bigwedge}
\def\bigor{\bigvee}
\def\msum{\oplus}
\def\mprod{\otimes}
\def\mod{\mathrm{mod}}
\newcommand{\iter}[1]{^{(#1)}}
\newcommand{\st}{\ensuremath{\quad\mathrm{s.t.}\quad}}
\newcommand{\norm}[1]{\ensuremath{\left\|#1\right\|}}
\newcommand{\support}[1]{\mathrm{supp}(#1)}
\newcommand{\rankf}[1]{\mathrm{rank}(#1)}
\def\rank{\mathrm{rank}}
\newcommand{\fun}[1]{\mathrm{#1}}
\newcommand{\abs}[1]{\ensuremath{\left|#1\right|}}
\newcommand{\setdef}[1]{\ensuremath{\left\{#1\right\}}}
\newcommand{\setspan}{\ensuremath{\mathrm{span}}}
\newcommand{\svec}[1]{_{[#1]}}
\newcommand{\row}[1]{_{#1:}}
\newcommand{\col}[1]{_{:#1}}
\newcommand{\havg}[1]{\langle\!\langle{#1}\rangle\!\rangle}
We begin this section by establishing the notation to be used throughout the paper.
Standard operations such as addition or subtraction are denoted as usual, $1+1=2$, $1-1=0$,
etc. Given two binary values $a$ and $b$, we use  $a \land b$ to denote their logical AND (Boolean product),
$a \lor b$ is their  OR (Boolean sum),  $a \msum b$ is modulo-2 addition (Boolean eXclusive OR), and
$\neg a$ is the 1's complement (Boolean negation) of $a$. The same notation is used for element-wise operations between vectors and matrices of the same dimension.

Let $x$ and $y$ be two binary vectors and $A$ and $B$ two binary matrices.
We denote the standard  inner and outer vector-vector, matrix-vector and matrix-matrix products as $x{\transp}y$, $xy\transp$, $Ax$ and $AB$.
The Boolean inner product between two vectors, $z = x\transp \bprod y$ is defined as $\bigor x_i \land y_i$. Similarly, the $C_{ij}$ element of the matrix product $C=A \bprod B$ is defined as the Boolean product between the $i$-th row of $A$, $A\row{i}$,  and the $j$-th column of $B$, $B\col{j}$. We define the modulo-2 inner product of $x$ and $y$ as $x \mprod y = (x+y)\;\mod \;2$. Finally, similarly to the Boolean case the $C_{ij}$ element of the modulo-2 product between two matrices, $C=A \mprod B$ is given by $A\row{i} \mprod B\col{j}$. 

The cardinality of a set $J$ is denoted by $|J|$. Given a set of indexes $J$, $B\col{J}$ denotes the
sub-matrix of columns of $B$ indexed by $J$, and $B\row{J}$ denotes a subset of its rows. These can
be combined with single indexes or other sets, e.g., $B_{iJ}$ contains the elements of row $i$ and
column indexes in $J$.  For a
vector $x$, its Hamming weight $h(x)$ is defined as the number of non-zero elements in $x$, that is
$h(x)=|\{i: x_i \neq 0\}|$. The same notation $h(A)$ is applied to count the non-zero elements of
matrices. The Hamming weight is usually referred to as the $\ell_0$ pseudo-norm,
$\|x\|_0=h(x)$. Note that, for binary vectors, $h(x)=\|x\|_0=\|x\|_1=\norm{x}_2^2$. Likewise, for
binary matrices, $\|A\|_0=\|A\|_{1,1}=\|A\|_F^2$. The function $\indicator(\cdot)$ is defined so
that $\indicator(cond)=1$ if $cond$ is true, and $0$ otherwise.
%Finally, we define the \emph{Hamming average} of a binary vector of length $m$ as
%$\havg{x} = \indicator(h(x) \geq m/2)$, that is, $1$ if the at least half of the elements of $x$
%are nonzero.

\begRevTwo
\subsection{A note on computational complexity}
\mnote{2.4}For each algorithm we provide a brief computational complexity analysis using a
simplified form of the familiar ``big O'' notation. Here a procedure which has order $O(m)$ is said
to have \emph{linear complexity}  in $m$, that is, it requires at most $am + b$ operations where $a$
and $b$ do not depend on $m$. Similarly $O(m^2)$ indicates \emph{quadratic complexity}, and
$O(m \log m)$ requires about $a(m \log m) + b$ operations in the worst case. This use is slightly
different than the formal Bachmann-Landau definition of $O(\cdot)$, which is defined in asymptotic
terms. For example, we might write $O(m \log m + p)$ as a shorthand to $O(m \log m) + O(p)$ to
indicate that a function requires about $am \log m + bp + c$ operations.
On the other hand, in all cases we will make a distinction as to the nature -- floating point,
integer, bitwise -- of the operations, as the constants involved in each case can vary greatly. For
example, as pointed out in the introduction, a bitwise inner product of two binary vectors of length
$256$ can computed in just \emph{two} CPU instructions, whereas it might require over $512$ instructions to compute the same operation on floating point or integer vectors.
\endRevTwo

\subsection{Dictionary Learning and sparsity}
\label{sec:background:dictionary-learning}
\begRevTwo

As described before, dictionary learning methods seek a to decompose $X$  as $X=DA+E$, where $D$ is such that we can achieve $\|E\col{j}\| \ll \|X\col{j}\|$ by using just a few atoms of $D$, that is, with a number of non-zero elements in $A\col{j}$ much smaller than $p$. The latter requirement is called \emph{sparsity}. Thus, dictionary learning is often also referred to as \emph{sparse modeling}, and finding $A\col{j}$ given $D$ as \emph{sparse coding}.
A typical approach to the dictionary Learning problem is to obtain a local solution by \emph{alternate minimization} in $D$ and $A$ of a cost function $f(D,A) + g(A)$,
\begin{eqnarray}
A\iter{t+1} =& \arg\min_{A} \{ f(D\iter{t},A) + g(A) \} \\
D\iter{t+1} =& \arg\min_{D} \{ f(D,A\iter{t+1}) + g(A\iter{t+1}) \},
\label{eq:dl}
\end{eqnarray}
where $f(\cdot)$ is a \emph{data fitting} term, usually $\|DA-X\|_F^2$ (here $\|X\|_F$ denotes the Frobenius norm of matrix $X$), and $g(\cdot)$  is a \emph{regularization} term which promotes sparsity in the columns of $A$. We now describe  the two  methods on which our methods are inspired.
\endRevTwo

\subsubsection{Method of Optimal Directions (MOD)} 
\label{sec:mod}
For the case $f(D,A)=\|DA-X\|_2^2$ and $g(A)=\sum_{j}\|A\col{j}\|_1$ the \emph{Method of Directions} (MOD)~\cite{mod}, is given by
\begin{eqnarray}
A_j\iter{t+1}\!\! &=&\!\! \arg\min_{a \in \reals^p} \{ \|x_j - D\iter{t}a \|_2^2 + \|a\|_1 \}, \\
D_r\iter{t+1}\!\! &=&\!\! U_r/\min\{1,\|U_r\|_2\},\nonumber\\
\!\!&&U=\!\!X(A\iter{t+1})\transp\left(A\iter{t+1}({A}\iter{t+1})\transp\right)^{-1},\,
\label{eq:mod}
\end{eqnarray}
The first step corresponds to an $\ell_1$-regularized least squares regression problem on each column of $A$, also known as LASSO~\cite{lasso}, a non-differentiable convex problem whose solution has been extensively studied in recent years, with several efficient algorithms designed specifically for the task (see e.g.~\cite{fista}).
In the second step, each atom $D\col{r}$ of the dictionary corresponds to a normalized down version of the least squares solution $u_r$.

The MOD algorithm is well suited for online dictionary adaptation, as both  $A{A}\transp$ and
$XA\transp$ can be efficiently updated when new columns are added to $X$. Furthermore, if new
samples arrive one at a time, the inverse of the Hessian matrix $({A}A\transp)^{-1}$ can be
efficiently updated via the Matrix Inversion Lemma~\cite{matrix-inv-lemma}. Moreover, as shown
in~\cite{online-dl}, excellent results can be still obtained if the Hessian is approximated by its
diagonal (in which case computing its inverse requires just $O(p)$ operations).
\begRevTwo
\paragraph*{Computational complexity} The offline version of the MOD dictionary \mnote{2.4}update step presented in Algorithm~\ref{eq:mod} requires $O(mnp) + O(p^2n) + O(p^3)$ floating point operations, which will generally be dominated by the $(p^2n)$ term. 
\endRevTwo   
\subsubsection{Matching Pursuit and the K-SVD algorithm} 
\label{sec:ksvd}

In this algorithm, proposed in~\cite{aharon06}, $f(E)=\norm{E}_F^2$ and $g(A)=h(A)$. The columns of $A$ are computed using a greedy method known as OMP (Orthogonal Matching Pursuit)~\cite{omp}, which   under certain conditions can be shown to provide the actual solution to the corresponding $\ell_0$-penalized least squares problem (see~\cite{tropp07}). A simpler variant of this step uses the (non-orthogonal) Matching Pursuit (MP)~\cite{mp}, which is described next in Algorithm~\ref{alg:mp}, 

\begin{algorithm}[ht] 
\caption{\label{alg:mp}Matching Pursuit}
\KwData{vector to encode $x$, dictionary $D$, maximum residual norm $\epsilon$, maximum coefficients weight $h_{\max}$}
\KwResult{Coefficients vector $a$}
Set iteration $t \leftarrow 0$, residual $r\iter{0} \leftarrow x$, initial coefficients $a\iter{0} \leftarrow 0$\;
Set $g\iter{0} \leftarrow D\transp{r\iter{0}}$, $G \leftarrow D\transp{D}$ \;
\While{$\norm{r\iter{t}} \geq \epsilon$ and $h(a) < h_{\max}$}{
  $i = \arg\max \left\{g\iter{t} \right\} $ \;
  ${\Delta} \leftarrow D\col{i}\transp r\iter{t}$ \;
  $a_i\iter{t+1} \leftarrow a_i\iter{t} + \Delta$ \;
  $r\iter{t+1} \leftarrow r\iter{t} - {\Delta}D\col{i} $\; 
  $g\iter{t+1} \leftarrow g\iter{t} - {\Delta}G\col{i} $\tcc*{(a)} 
  $t \leftarrow t+1 $ \;
}
\Return $a \leftarrow a\iter{t}$ \;
\end{algorithm}

What MP does at each iteration is to project the residual onto the atom that is most correlated to it, and then remove the projection from the residual. For this to work well, the atoms must be normalized to have $\ell_2$ norm 1.
The vector $g$ keeps track of the correlation between the residual $r$ and the dictionary $D$.  Its update $(a)$ is derived as follows: 
\begin{eqnarray}
g\iter{t+1}
 &=& D\transp{r\iter{t+1}}=D\transp(r\iter{t}-{\Delta}D_i)\nonumber\\
 &=& g\iter{t}-{\Delta}D\transp{D}\col{i} =g\iter{t}-{\Delta}G\col{i}.
\label{eq:corr-up}
\end{eqnarray}
Note that, by means of \refeq{eq:corr-up}, updating $g$ requires only $O(p)$ floating point operations, whereas the na\"{\i}ve update	 requires $O(mp)$ operations. This same trick, with a few modifications, will also be useful in the binary case.
\begRevTwo \mnote{2.4}As the Gramm matrix is computed only once, the overall complexity of MP for processing all $n$ samples is $O(p^2m + p + kmn)$ floating point operations.\endRevTwo

The dictionary update of K-SVD performs a simultaneous update of each atom $D\col{r}$ and the row of $A$ associated to it. This update is described in Algorithm~\ref{alg:ksvd}. 
\begin{algorithm}
\KwData{Current iterate $(D\iter{t}$, $A\iter{t})$}
\KwResult{Next iterate $(D\iter{t+1}$,$A\iter{t+1}$}
\For{$r=1,\ldots,p$}{
  $J \leftarrow \{j: A_{rj}\iter{t} \neq 0\}$ \;
  $R \leftarrow X\col{J} - D\iter{t}A\col{J}\iter{t} + D\col{r}\iter{t}(A_{rJ}\iter{t})$ \;
  $U{\Sigma}V\transp \leftarrow \mathrm{SVD}(R)$ \;
  $D\col{r}\iter{t+1}\leftarrow U\col{1}$ \;
  $A_{rJ}\iter{t+1} \leftarrow V\col{1}$ \;
}
\caption{\label{alg:ksvd}K-SVD Dictionary update.}
\end{algorithm}
\begRevTwo
\paragraph*{Computational complexity} Each of the $p$ updates of $D$ requires $O(mn)$
\mnote{2.4}   operations for updating the residual plus another $O(n^2)$ operations for computing the first
  pair of singular eigenvectors. This gives a total $O(pn^2) + O(pmn)$ operations, which in general
  will in our case ($n \gg m \approx p$) will be significantly more expensive than the $O(p^2n) + O(pmn) + O(p^3)$ complexity of MOD. 
\endRevTwo
\section{Binary Dictionary Learning}
\label{sec:bdl}

Below we describe our dictionary learning methods, which assume a given fixed dictionary size $p$
and employ the traditional alternate descent approach to obtain the best model $(D,A,E)$ for that
$p$. We leave the description of the top-level model selection algorithm for choosing the best model
order $p$ to Section~\ref{sec:model-selection}.

Both BMF algorithms share a common coefficients update step, the Binary Matching Pursuit (BMP)
algorithm, and  two choices for the dictionary update step: MOB (a binarized version of MOD) and
K-PROX (combining ideas of K-SVD and Proximus); these are detailed next.

\subsection{Coefficients update via Binary Matching Pursuit (BMP)}
\label{sec:bdl:bmp}

\begin{algorithm}
\KwData{sample to encode $x$, dictionary $D$, initial coefficients $a\iter{0}$, maximum coeffs. weight  $h_{\max}$, maximum residual weight $w_{\max}$.}
\KwResult{Optimum coefficients for $x$, $a$}
Set iteration $t=0$, coefficients $a\iter{0}=a_0$ ,residual $r\iter{0}=x \oplus D{\mprod}a\iter{0}$\;
Set modulo-2 Gramm matrix $G \leftarrow D\transp \mprod D$\tcc*{(a)}
Set residual correlation $g\iter{0} \leftarrow D\transp{r\iter{0}}$\tcc*{(b)}
\While{$h(r\iter{t}) \geq w_{\max}$ \textbf{and} $t < h_{\max}$ }{
  $k \leftarrow \arg \max_l \{\;|g\iter{t}_l|\;/\;\|D\col{l}\|_0\;\} $ \tcc*{(c)}
  \If{$g\iter{t}_{k} = 0$} { 
  	  \Return $a \leftarrow a\iter{t}$ \; 
  }  
  $r\iter{t+1} \leftarrow r\iter{t} \oplus  D\col{k} $\; 
  \If { $h(r\iter{t+1}) \geq h(r\iter{t})$ } 
  {
	  \Return $a \leftarrow a\iter{t}$ \;
  }
  \eIf {$a\iter{t+1}_{k} = 1$} {
    $a\iter{t+1}_{k} \leftarrow 0$;  $g\iter{t+1} \leftarrow g\iter{t} - G\col{k}$\tcc*{(d)}
    } {
    $a\iter{t+1}_{k} \leftarrow 1$;  $g\iter{t+1} \leftarrow g\iter{t} + G\col{k}$ \tcc*{(d')}
    }
} 
\Return $a \leftarrow a\iter{t}$ \;
\caption{\label{alg:bmp} Binary Matching Pursuit.}
\end{algorithm}

In essence, BMP is a binarized version of the Matching Pursuit Algorithm~\ref{alg:mp}. For a given
sample $x$, we begin ($t=0$) with an initial coefficients vector $a\iter{0}=a_0$, a residual
$r\iter{0}_j=x \oplus D \mprod a\iter{0}$ and an initial vector $g\iter{0} = D\transp r\iter{0}$
which keeps track of the correlation between the columns of $D$ and $r\iter{t}$. Then, at each
iteration $t$ we determine the atom $D\col{k}$ which is most correlated to $r\iter{t}$. We
then \emph{toggle} the coefficient corresponding to that atom, $a_k\iter{t}$, and update
$r\iter{t}$ and $g\iter{t}$ accordingly. The pseudocode is given in Algorithm~\ref{alg:bmp}. 
Some steps of this algorithm, marked as $(a)$, $(b)$, $(c)$ and $(d)$ (appearing twice) in
Algorithm~\ref{alg:bmp}, are not obvious from the overall description of the algorithm and need to
be clarified. In $(a)$, the \emph{modulo-2 Gramm matrix} of $D$ is computed; this matrix is used in
an analogous way in Algorithm~\ref{alg:mp} for the fast update of the correlations vector $g$ as
described in \refeq{eq:corr-up}. In $(b)$, we compute the standard correlation between the columns of
$D$ and $r$. In $(c)$, since the atoms are not normalized,  the best candidate is chosen using a
form of normalized correlation called \emph{association accuracy}~\cite{association-accuracy},
$g\iter{t}_l / \|D\col{l}\|_2^2 = g\iter{t}_l / \|D\col{l}\|_0$. 

Finally, in $(d)$ and $(d')$, despite the correlation vector $g$ being initially computed using
the \emph{standard} matrix-vector product, its updated has exactly the same form
as \refeq{eq:corr-up}, but the Gramm matrix $G$ is actually computed using the modulo-2; 
$(d)$ corresponds to the case when $\Delta=1$, that is, when $a_k$ is switched from $0$ to $1 $, and
$(d')$ corresponds to $\Delta=-1$, when $a_k$ is switched off. (This curious result is easily verified by writing down the corresponding arithmetic.)
\begRevTwo
\paragraph*{Computational complexity} The initialization of BMP requires $O(p^2m)$ binary
operations for computing the Gramm matrix and integer ones $O(p^2m)$ for the \mnote{2.4}correlation vector
$g$. Then, for each sample $X\col{j}$ a maximum of $h_{\max}$ iterations can be required, each one
requiring $O(p)$ floating point operations for finding the most correlated atom, another $O(p)$
integer operations for updating the residual $r$, and another $O(p)$ integer ones for updating $g$, for a total of $O(h_{\max}p)$ operations per sample. Overall, a whole pass over the $n$ data samples requires $O(p^2m)+ O(p) + O(h_{\max}pn)$ operations.
\endRevTwo
\subsection{MOB: Method Of Binary Directions}
\label{sec:bdl:mod}
 
Here we want to update $D$ so as to minimize the Hamming weight $h(E)$ of the residual matrix $E =
 X \oplus D \mprod A$. Suppose we want to update the $r$-th atom at iteration $k$. The
 affected columns will only be those for which the coefficients in $A$ corresponding to that atom
 are non-zero. We define $J=\{j : A_{rj} \neq 0 \}$ to be the set of indexes of those columns. What
 we want is to update $D\col{r}$ so that  the weight of the columns of the residual 
affected by it is minimized,
 \begin{eqnarray}
 D\col{r}\iter{t+1}  &=& \arg\min_{d \in \{0,1\}^m} \sum_{j \in J}  h(E\iter{t}\col{j} \oplus d) \nonumber\\
 &=& \arg\min_d \sum_{j \in J} (\sum_i E\iter{t}_{ij} + |J| d_i).
\label{eq:mob1}
 \end{eqnarray}

According to~\refeq{eq:mob1}, the optimization of $D\col{r}\iter{t+1}$ is separable in each of its elements,
 \begin{eqnarray}
 D_{ir}\iter{t+1}  &=& \arg\min_{u \in \{0,1\}} \sum_j E_{ij}\iter{t} \oplus u \nonumber\\
 &=& \arg\min_{u \in \{0,1\}} \left|\;\sum_{j\in J} \;E_{ij}\iter{t} - |J|u\;\right| \nonumber\\
 &=& \arg\min_{u \in \{0,1\}} |\,h( E_{iJ}\iter{t} ) - |J|u\,|\nonumber\\
 &=& \indicator\left( \frac{ h( E_{iJ}\iter{t} ) }{|J|} > \frac{1}{2} \right)
\label{eq:mob2}
 \end{eqnarray}
Note that when $h(E_{iJ})/|J| = 1/2$ both $0$ and $1$ are optimum, in which case we use $0$. Also, note that \refeq{eq:mob2} can be rewritten as
$$D_{ir}\iter{t+1} = \indicator\left( \frac{h(EA\row{r}\transp)} {h(A\row{r})} > \frac{1}{2} \right).$$
As these statistics can be easily updated as new samples arrive, it follows that MOB is suited for fast online dictionary adaptation. (Our implementation of this feature is still work in progress.)
\begRevTwo
\paragraph*{Computational complexity of MOB} Our current (offline) \mnote{2.4} implementation requires $O(mnp)$ bitwise operations for the modulo-2 product $E A\transp$, $O(np)$ integer operations for computing the weights of the rows of $A$, and $O(mp)$ integer comparisons for updating $D$, for a total of $O(mnp)$ binary plus $O(mp)$ integer operations.
\endRevTwo
\subsection{K-PROX: Dictionary Update via Proximus}
\label{sec:bdl:k-prox}

In this case, following the K-SVD concept, we want to obtain the best rank-one approximation to the
residual $E$ obtained after removing the contribution of $D\col{r}$. Let $J = \{j: A_{rj} \neq 0 \}$
and  $R_{J}=D\col{r}{\mprod}A_{rJ} \oplus E_{J}$. We then have,
\begin{equation}
(D\col{r}\iter{t+1},A_{rJ}\iter{t+1}) = \arg\min_{u,v} h\left( R\col{J}\iter{t} \oplus uv\transp \right).
\label{eq:bsvd1}
\end{equation}
As the name K-PROX implies, our approximation to the NP-hard problem \refeq{eq:bsvd1} is based on the Proximus algorithm~\cite{proximus}, summarized in Algorithm~\ref{alg:proximus},
\begin{algorithm} 
\caption{\label{alg:proximus} Proximus}
\KwData{matrix $X \in \{0,1\}^{m{\times}n}$, $u\iter{0} \in \{0,1\}^m$, $v\iter{0} \in \{0,1\}^n$ }
\KwResult{Vectors $u$, $v$ so that $X \approx uv\transp$}
Set iteration $k=0$\;
\Repeat {$u\iter{t+1}(v\iter{t+1})\transp = u\iter{t}(v\iter{t})\transp$} {
  $u\iter{t+1}_i\!\! \leftarrow \indicator \left( X\row{i} v\iter{t} > h(v\iter{t})/2 \right),\;i=1,\ldots,m$ \;
  $v\iter{t+1}_j\!\! \leftarrow \indicator\left( X\col{j}\transp{u\iter{t+1}}\!>\!h(u\iter{t+1})/2 \right),j=1,\ldots,n$ \;
  $k \leftarrow k+1 $ \;
}
\end{algorithm}
Interestingly, Algorithm~\ref{alg:proximus} provides a local optimum to the rank-one approximation that we seek. This is stated in Proposition~1 below.
\begin{proposition}
The output $(u,v)$ of the Proximus Algorithm~\ref{alg:proximus} is a local optimum of the problem $\min \|X - uv\transp\|_0$. 
\end{proposition}
\begin{proof}
 Given $v\iter{t}$, it is easy to check that the update $u\iter{t+1}$ in Algorithm~\ref{alg:proximus} is the value of $u$ that \emph{globally} minimizes $\|X \oplus uv\iter{t+1} \|$ (if $ s\iter{t}_i = w\iter{t}/2$, both $0$ and $1$ are equally optima; in such case, we default to $0$). The same happens with the update $v\iter{t+1}$ given $u\iter{t+1}$. Therefore, $h(E\iter{t})=h(X \oplus u\iter{t}(v\iter{t})\transp)$ cannot increase with $k$. As $h(E\iter{t}) \geq 0$ is bounded, non-increasing, and the iterates can take on a finite number of values, the sequence $h(E\iter{t})$ must converge  after a finite number of steps. Let $(u,v)$ be the arguments at which the stopping condition is satisfied. By definition of the algorithm, no change in $u$ or $v$ decreases the objective. This guarantees that $(u,v)$ is a local minimum in a Hamming ball of radius at least $1$.\footnote{We cannot guarantee that a simultaneous change in a single coordinate of $u$ and a single coordinate of $v$ will not decrease the cost function!.} 
\end{proof}
\begRevTwo
\paragraph*{Computational complexity of K-PROX} As with the K-SVD algorithm, we update $D$ one atom at a time.  This requires $O(mn)$ operations for computing \mnote{2.4}the residual in \refeq{eq:bsvd1} and running Algorithm~\ref{alg:proximus}, which takes a finite number of iterations requiring $O(mn) + O(np)$ bitwise operations and the same number of integer comparisons. 
\endRevTwo

\subsection{Initialization}
\label{sec:bdl:init}

Initialization is of paramount importance to the success of any non-convex matrix factorization method. At the same time, there is no provably optimum way of doing so, otherwise we would be contravening the NP-hard nature of the factorization problem itself. We are left with heuristics based on intuition and prior information, if any. Ultimately, \emph{initialization is an art}, and also an engineering decision which may depend on several aspects. As an example, we could use the resulting factorization obtained with \emph{any} of the existing methods mentioned in Section~\ref{sec:intro} as an initial point. In this case, to maintain scalability and simplicity, we experiment with two  simple but generally effective methods drawn also from dictionary learning literature: given dictionary size $p$, the first draws $p$ atoms using a pseudo-random Bernoulli$(\theta)$ distribution; we use $\theta=1/2$, but other values could be used to reflect prior information on the problem.
The second method is to draw $p$ columns from $X$ at random and use them as the initial atoms. WE report on both strategies on Section~\ref{sec:results}.

\section{Model selection}
\label{sec:model-selection}

Beyond theoretical results and simulations, when confronted with real data,
the true underlying model governing the generation of data is rarely
available. In such situations, models can only be assumed to be tools for
understanding the data at hand, and the best choice is dictated by how much
regularity or structure each candidate model is capable to grasp from that
data. In particular, the complexity of a model is limited by the amount of
data available to estimate the different parameters of the model. An overly
complex model will tend to \emph{overfit} the data, whereas an overly simple
one will miss important details. The problem of model selection is that of
finding the best model for a given data set. Typically, this is done by
seeking a balance between the \emph{goodness of fit} of a model, usually
expressed in terms of the likelihood of the data given the model, and a
measure of the \emph{model complexity}; some popular examples in this
category are the Bayesian Information Criterion (BIC)~\cite{bic}, the
Akaike's Information Criterion~\cite{aic}, and the Minimum Description
Length (MDL) principle~\cite{mdl1,mdl2,mdl3}.

MDL translates the model
selection problem as one of data compression, where both the data and the
model have to be (hypothetically) transmitted and perfectly recovered using
some encoding mechanism. The tension between complexity is then resolved by
the number of bits (\emph{codelength}) required to describe the data in
terms of the model (\emph{stochastic complexity}) and the number of bits
required to describe the model itself (\emph{model complexity}). The
original version of MDL~\cite{mdl1} made this division explicit, and is
asymptotically equivalent to BIC. However, the modern version of
MDL~\cite{mdl2,mdl3} uses the more recent
information-theoretic \emph{universal coding theory} to make the
aforementioned balance implicit by producing an optimum, joint description
of both the model and the data which is furthermore independent of arbitrary
choices of, for example, the way the model is parameterized.
These advantages make MDL an appealing method for model selection.

In our case, the data $X$ is described by the triplet $(D,A,E)$. We describe
each of these components separately using universal codes, so that
$L(X)=L(D)+L(A)+L(E)$ is the total codelength of describing $X$. Here the
tension between a good fit and a simple model is represented respectively by
$L(E)$ and $L(D)+L(A)$.

\subsection{\revTwo{Forward selection algorithm}}
\begRevTwo
 Model selection is usually formulated as two nested problems. The inner
problem is how to search for the best model $(D,A,E)$ among all models of
\mnote{2.5} order $p$. The second problem is to choose the best model for $X$ among all
possible models of order $p \geq 0$.  Given a codelength function $L(X)$,
our method uses a \emph{forward selection} strategy to sweep over all
models. Starting with a initial order $p=p_0$, we approximate the best model
given $p$ using one of our dictionary learning algorithms, and then
gradually increase $p$ until the codelength $L(X)$ is no longer diminished.

\mnote{2.4}
 In going from $p$ to $p+1$ we employ a \emph{warm restart} strategy, that is, atoms and coefficients  learned for model order $p$ are used as the starting point for learning the $p+1$-th order model. The overall procedure is summarized in Algorithm~\ref{alg:fwd},

\begin{algorithm} 
\caption{\label{alg:fwd}MDL-based Forward Selection Algorithm}
\KwData{matrix $X \in \{0,1\}^{m{\times}n}$, initial order $p_0$ and model $(D\iter{0},A\iter{0},E\iter{0})$ }
\KwResult{Selected model $(D^*,A^*,E^*)$}
$p \leftarrow p_0$ \; $(D\iter{p},A\iter{p},E\iter{p}) \leftarrow
(D\iter{0},A\iter{0},E\iter{0})$ \; $L\iter{p} \leftarrow
L(D\iter{p})+L(A\iter{p})+L(E\iter{p})$ \;
\Repeat {$L\iter{p} \geq L\iter{p-1}$} {
Initialize the rank-$1$ model $(d,a)$ using $E\iter{p}$ as input \;
$(\tilde{D},\tilde{A},\tilde{E}) \leftarrow 
([D\iter{p}|d],\,[(A\iter{p})\transp|a\transp]\transp\,,\,E\iter{p}-da\transp) $ \;
Adapt $(D\iter{p+1},A\iter{p+1},E\iter{p+1})$ using $(\tilde{D},\tilde{A},\tilde{E})$ as the starting point\;
$L\iter{p+1} \leftarrow L(D\iter{p+1})+L(A\iter{p+1})+L(E\iter{p+1})$ \;
$p \leftarrow p+1$ \;
}
$(D^*,A^*,E^*) \leftarrow (D\iter{p-1},A\iter{p-1},E\iter{p-1})$ \;
\end{algorithm}
\endRevTwo

\subsection{Codelength computation}

It remains to describe how we compute $L(X)$. The universal compression of
binary sources has been extensively studied in the literature. Moreover, we
do not need to perform a real encoding; we need only to compute the
codelength. One particularly simple method, for which the codelength is easy
to compute, is \emph{enumerative coding}~\cite{enum}. Given a binary string
$x$ of length $n$ and $r=h(x)$, its enumerative code is
composed of two parts. The first one describes $r$ with
$\lceil\log_2(n)\rceil$ bits, and the second one describes the index of $x$
in the lexicographically ordered list of all binary strings of length $n$ and weight $r$, which requires $\lceil\log_2{r \choose n}\rceil$ bits. The
total codelength is then
\begin{equation}
L(x) = \lceil\,\log_2(n)\,\rceil + \left\lceil\log_2{r \choose n}\right\rceil.
\label{eq:codelength}
\end{equation}
(Note that $\log{r \choose n}$ can be accurately approximated by using
Stirling's formula, i.e., requiring $O(1)$ operations.) As prior
information, we expect the different columns of $D$ to represent particular
patterns in the data, the corresponding rows of $A$ their appearance in $X$,
and the different rows (corresponding to different variables of the data
samples) of $E$ to exhibit different patterns also. Accordingly, we encode
each column of $D$ and each row of $E$ and $A$ with its own code,
\begin{equation}
L(X) = \sum_{i=1}^m L(E\row{i}) + \sum_{k=1}^p L(D\col{k}) + \sum_{k=1}^p L(A\row k) 
\label{eq:model-codelength}
\end{equation} 
\begRevTwo
\paragraph*{Computational complexity} The cost of each forward selection step is \mnote{2.4-5}clearly dominated by the dictionary learning algorithms. The codelength evaluation is negligible, requiring  $O(mn) + O(mp) + O(np)$ operations to count the number of zeros in each component $(D,E,A)$. 
\endRevTwo

\section{Results and discussion}
\label{sec:results}
\begin{figure*}[t]
\centering%
\includegraphics[height=1.8in]{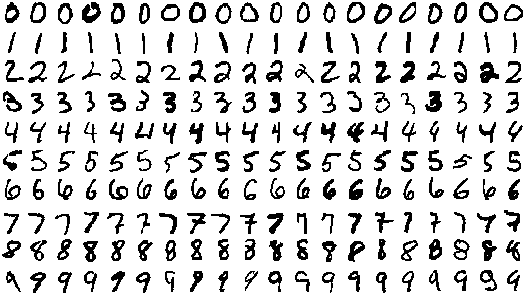} %
\includegraphics[height=1.8in]{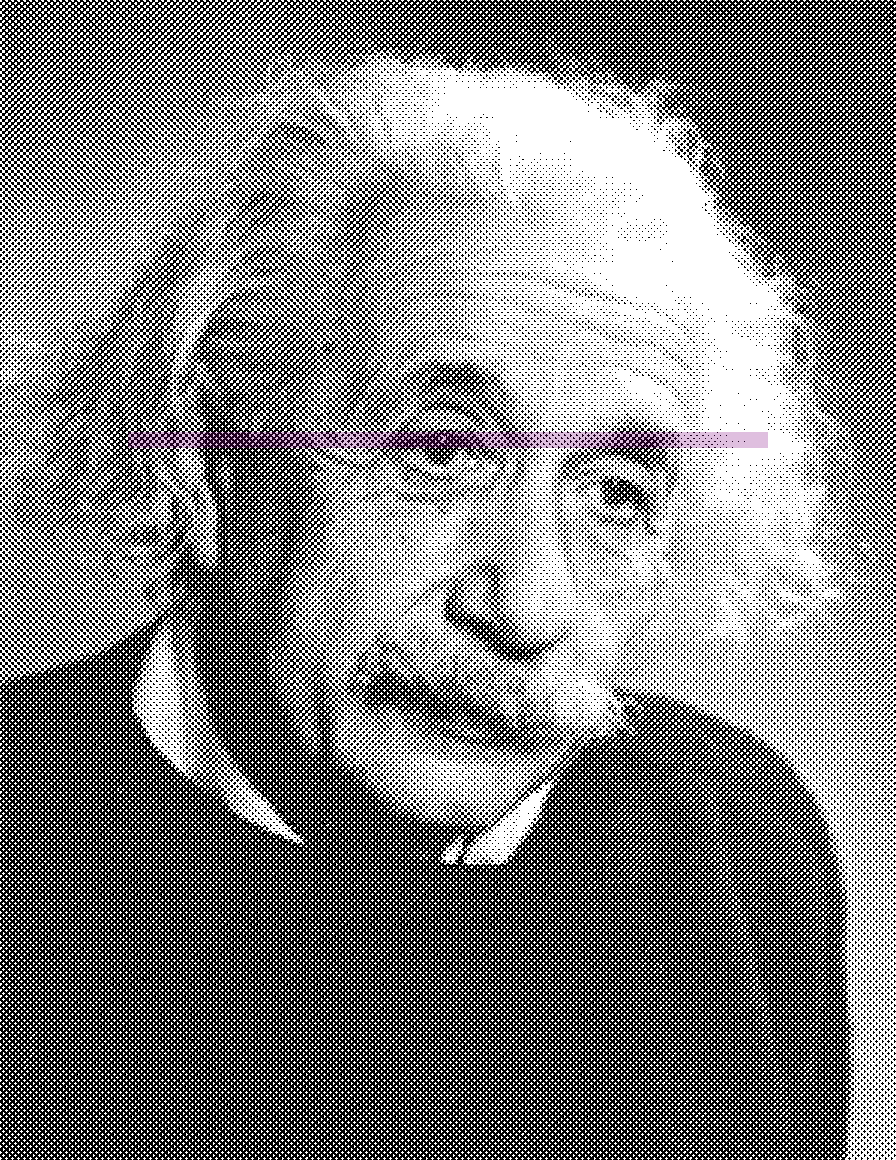} %
\includegraphics[height=1.8in]{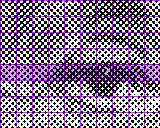}
\caption{\label{fig:datasets} \begRevTwo Left to right: a few samples of the MNIST dataset; halftone image of Einstein with the stripe used in Figure~\ref{fig:einstein-kprox} highlighted in magenta; detail of Einstein's left eye and the $16{\times}16$ blocks partition; including  part of the stripe.  the first case, we expect the atoms in the final dictionary to resemble numbers of different shapes (see e.g.~\cite{cvpr10}). In the case of Einstein, we expect the dictionary atoms to resemble the halftoning patterns observable in the $16{\times}16$ blocks shown on the right picture.\endRevTwo}
\end{figure*}

\begRevTwo
The main objectives of the following experiments are three. The first is to
demonstrate the interpretability of the resulting models; we expect to
\mnote{1.2}\mnote{2.6} obtain atoms which exhibit recognizable patterns of the input data. The
second is to see the effect of different initialization strategies on the
final result. The third is to study the numerical properties of the methods, in particular their convergence rate and computational cost.\footnote{The version used in this
paper can be downloaded from \url{http://iie.fing.edu.uy/~nacho/bmf/bmf.zip}; this
includes scripts and data to reproduce all the results shown in this section
and many more not discussed in this paper for lack of space. The latest
version is available as a GIT
repository \url{https://gitlab.fing.edu.uy/nacho/bmf}.}
\endRevTwo
The first dataset is a binarized version of MNIST~\cite{mnist}, a set of  $n=10000$ $17{\times}17$ images of handwritten digits;
here each columns of $X$ contains the vectorized version of a digit
($m=289$). The second consists of all the $m=4088$ non-overlapping
blocks of size $16{\times}16$ of the \emph{halftone Einstein}, a
binary $1160{\times}896$ image; the columns of $X$ are the $4088$
vectorized blocks ($m=256$) of the image.  As can be seen in
Figure~\ref{fig:datasets}, both datasets have easily recognizable patterns.
\begin{figure*}[t]
\centering%
\includegraphics[width=\textwidth]{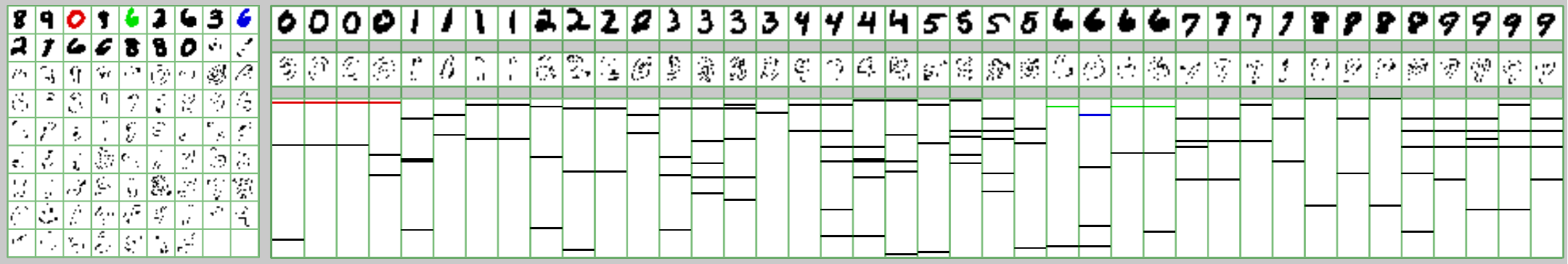}%
\caption{\label{fig:mnist-mob}\begRevTwo MNIST model obtained using Forward selection with MOB at each step. The model was initialized with $p_0=16$ random samples. The final dictionary, with $p=79$ atoms, is shown to the left  as a mosaic from top to bottom and left to right. The three rows to the right of the dictionary show $40$ samples from $X$ (top), their final residual from $E$ (middle), and the corresponding  coefficients from $A$ (bottom). Each column of $A$ is represented as a vertical pattern of $79$ bands: the top band corresponds to the first atom, and the bottom one to the 79th;  a black band on the $i$-th row indicates that atom $i$ is being used to represent the sample on top of the same column, giving rise to the residual shown in the middle. %
In general, coefficients are not easy to interpret. However, in some cases the correspondence is clear. In this case the four digits ``0'' use the third atom (painted in red). Something similar happens with the ``6''s: three of them use the 5th atom (green) and one uses the 9th (blue). The corresponding coefficients are marked as red, green and blue bands in the bottom-right picture.
\endRevTwo}
\end{figure*}

\begin{figure*}[t]
\centering%
\includegraphics[width=1.0\textwidth]{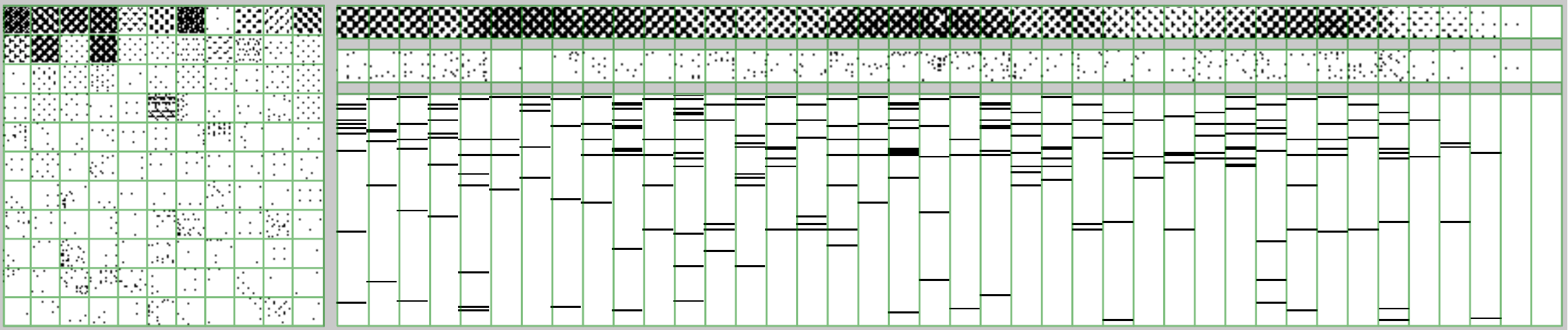}
\caption{\label{fig:einstein-kprox}\begRevTwo Einstein model obtained using forward selection and K-PROX. Here too we used random samples for the initialization. As in Figure~\ref{fig:mnist-mob}, we show the resulting dictionary on the left, and the samples (top), residuals (middle) and coefficients (bottom) for the patches contained in the stripe highlighted in Figure~\ref{fig:datasets}. \endRevTwo }
\end{figure*}
\begin{figure}
\centering%
\includegraphics[height=1.0in]{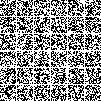} % OK
\includegraphics[height=1.0in]{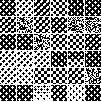} % OK
\includegraphics[height=1.0in]{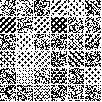}\\[1ex]% OK
\includegraphics[height=1.0in]{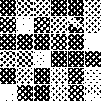} % OK
\includegraphics[height=1.0in]{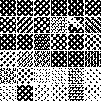} % OK
\includegraphics[height=1.0in]{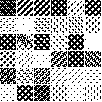}\\[1ex]% OK
\caption{\label{fig:init-einstein} \begRevTwo Effect of initialization on Einstein. \begRevTwo Top row: initial dictionary drawn from a Bernoulli($1/2$) process (left), resulting MOB (center) and K-PROX (right) models for $p=36$. Bottom row: initial dictionary using random columns from $X$ (left), resulting MOB (center) and  K-PROX (right) dictionaries. Both initialization methods worked well in this case using both MOB and K-PROX. In all cases, the  dictionaries evolved so that some atoms resemble halftoning patterns. Other atoms were not changed, indicating that they were never used. 
\endRevTwo}
\end{figure}
\begin{figure}[t]
\centering%
\includegraphics[height=1.0in]{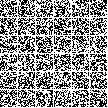} % OK
\includegraphics[height=1.0in]{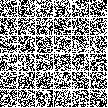} % OK
\includegraphics[height=1.0in]{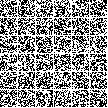}\\[1ex]% OK
\includegraphics[height=1.0in]{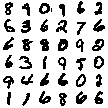} % OK
\includegraphics[height=1.0in]{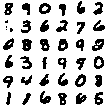} % OK
\includegraphics[height=1.0in]{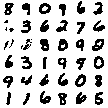}\\[1ex]% OK
\caption{\label{fig:init-mnist} \begRevTwo Effect of initialization on MNIST. Top row: initial pseudo-random dictionary using a Bernoulli($1/2$) model (left), result of MOB (center) and of K-PROX (right) for $p=36$. Bottom row: initial dictionary using random samples from the dataset (left),  MOB (center) and K-PROX (right) dictionaries. Here  the pseudo-random initialization does not work; no atom is ever used, and so both algorithms stop at iteration $1$. In the case of the samples-based initialization, both final models show some adaptation. \endRevTwo}
\end{figure}
\begRevTwo
\subsection{Interpretability of the resulting models}

Figure~\ref{fig:mnist-mob} shows the MNIST model obtained using MOB and forward selection, The dictionary and a few columns from the coefficients and residual \mnote{1.3}\mnote{2.6}matrices $A$ and $E$; atoms from $D$ and samples from $E$ are represented as mosaics with each sample/atom displayed as a tile. As can be seen, many atoms look like numbers; other atoms exhibit number-like silhouettes. The results obtained with K-PROX (shown in the supplementary material) are very similar in all aspects. Figure~\ref{fig:einstein-kprox} shows the model obtained for the Einstein image using K-PROX; we show the dictionary, a few samples; again, the results are clearly interpretable in terms of the patterns that can be observed in the data. In this case too the results obtained with MOB (in the supplementary material) are very similar.

\subsection{Sensitivity to initialization}

Figures~\ref{fig:init-einstein} and~\ref{fig:init-mnist} show initial and final dictionaries of fixed size $p=36$ for MNIST and Einstein, using both dictionary learning methods, but using each of the two initialization methods described in Section~\ref{sec:bdl:init}. In this case, the random samples initialization does a good job in both cases, whereas the pseudo-random initialization fails miserably on the MNIST case. These results should be taken with a grain of salt, only to show how different initialization methods can work (or fail) under different circumstances.

\subsection{Numerical results}
\begin{figure}
\centering%
\includegraphics[width=0.235\textwidth]{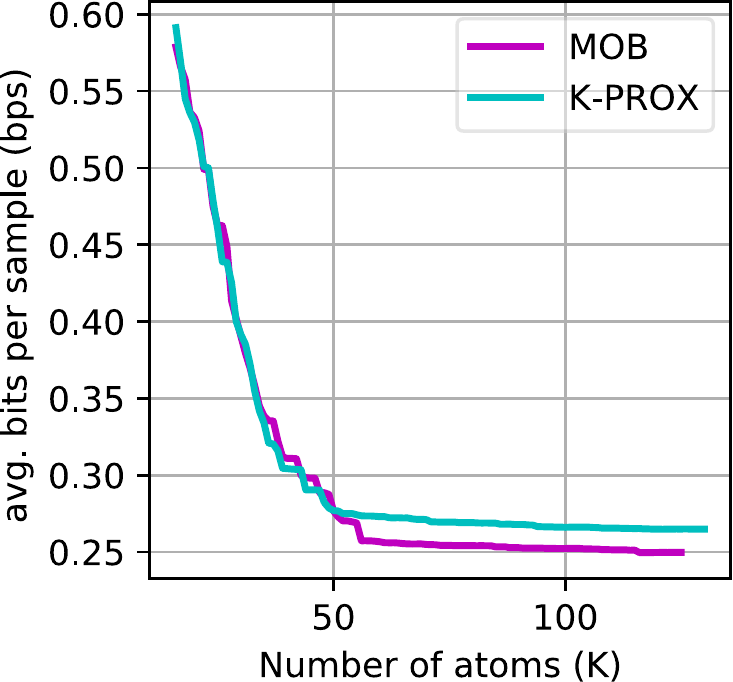}\hspace{2ex}%
\includegraphics[width=0.23\textwidth]{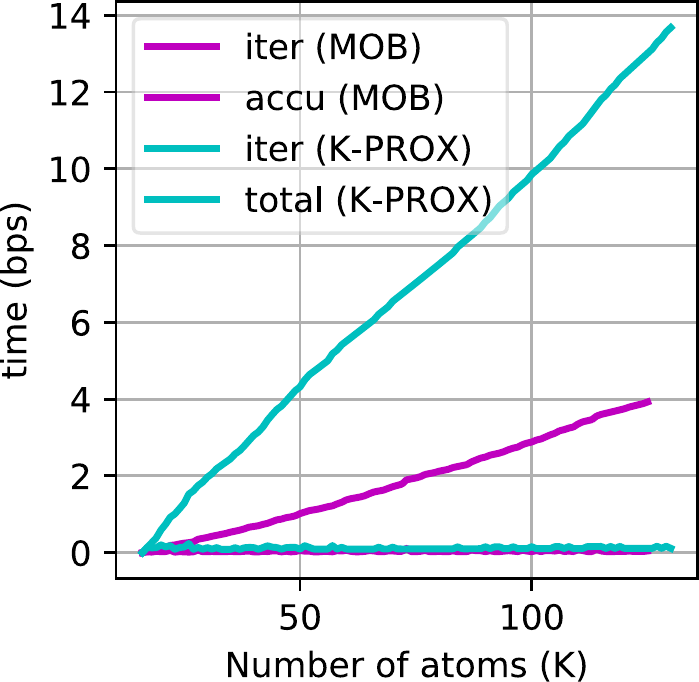}\\[1.5ex]
\includegraphics[width=0.23\textwidth]{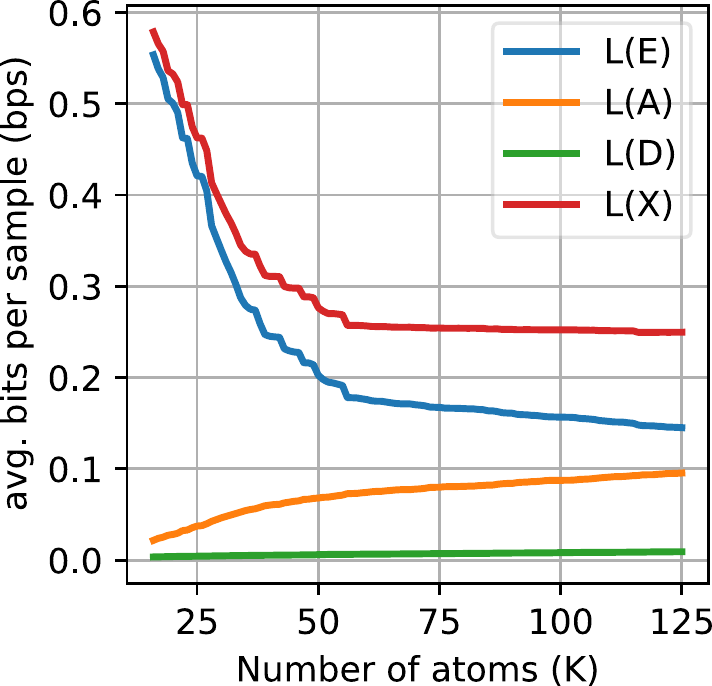}\hspace{2ex}%
\includegraphics[width=0.23\textwidth]{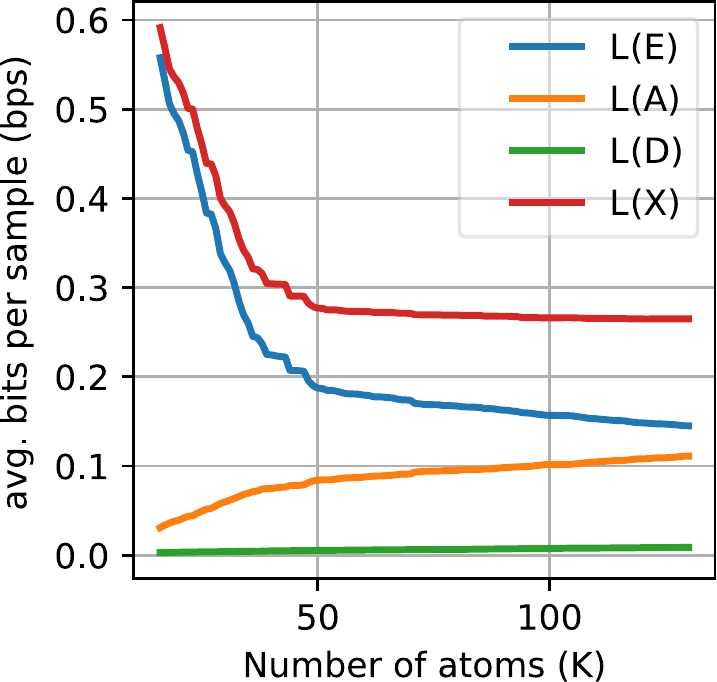}%
\caption{\label{fig:numerical-results-atoms}
\begRevTwo%
Convergence of Forward Selection on Einstein. Top to bottom, left to right:  MDL cost function (in avg. bits per sample) for MOB and K-PROX; computational cost (in seconds) for both variants; break-down of the cost function in its three parts ($E$, $A$ and $D$) for MOB; same for K-PROX. %
Both MOD and K-PROX produced similar models in the end. K-PROX, however, required  more time to run ($14$s) than MOB ($4$s). Thanks to the warm-restart strategy, the cost per forward selection step is small ($0.11$s for MOB and $0.03$s for K-PROX) and approximately constant despite the growing model size $p$. Finally, both break-downs show typical MDL curves: as $p$ increases, the stochastic complexity ($L(E)$) decreases while the model cost $L(A)+L(D)$ increases.\endRevTwo}
\end{figure}
\begin{figure}
\centering%
\includegraphics[width=0.23\textwidth]{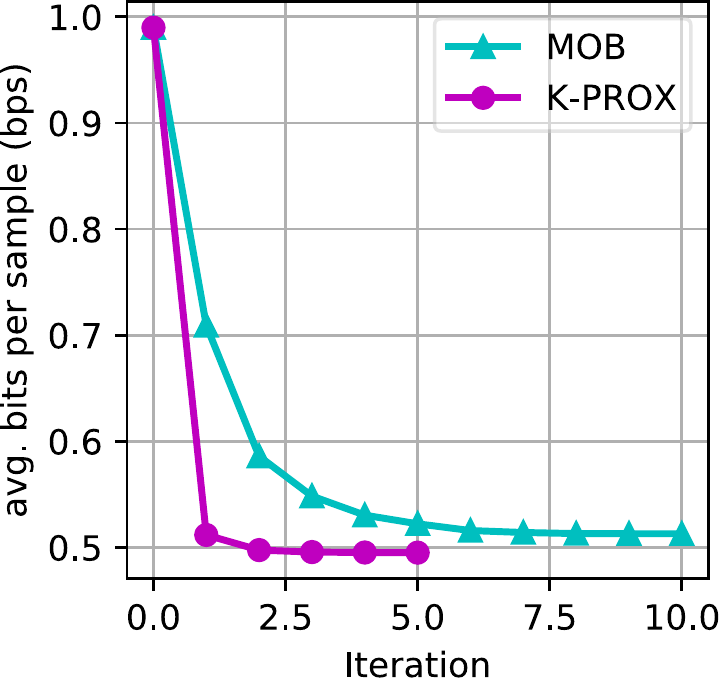}\hspace{2ex}%
\includegraphics[width=0.23\textwidth]{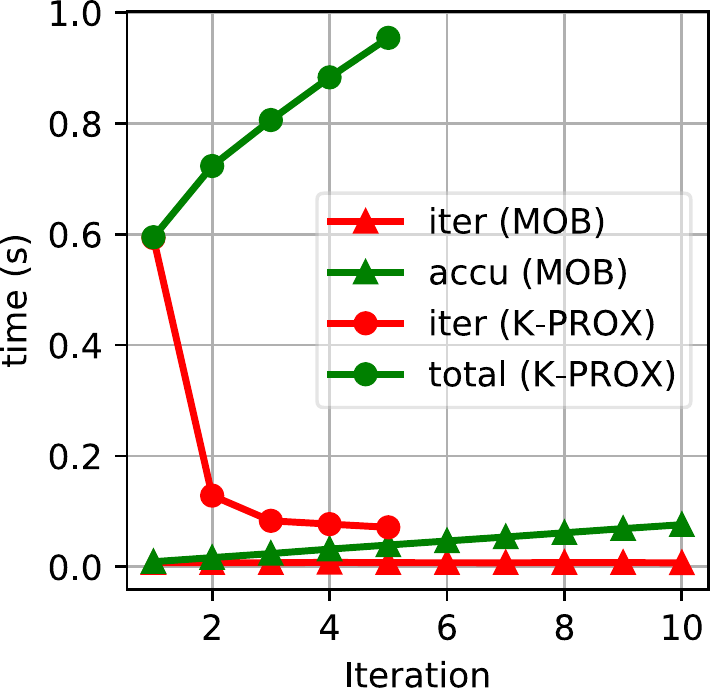}\\[1.5ex]
\includegraphics[width=0.23\textwidth]{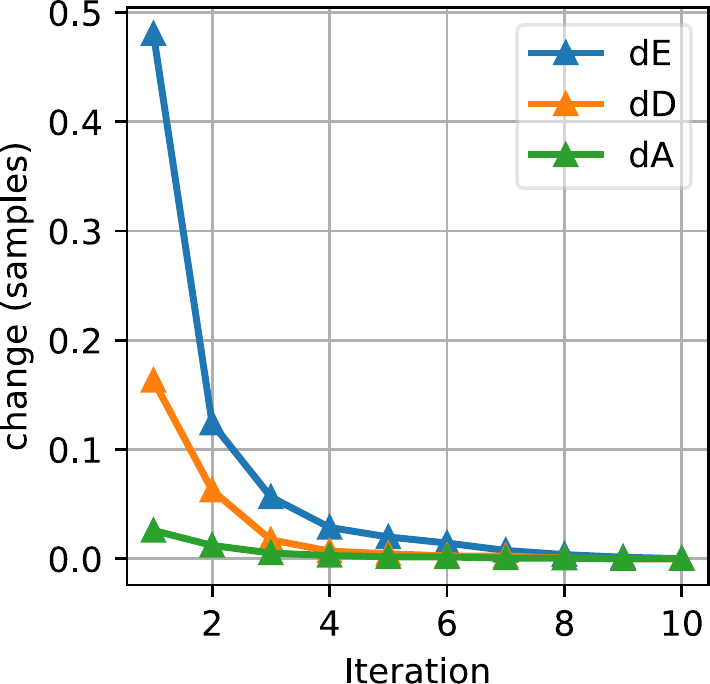}\hspace{2ex}%
\includegraphics[width=0.23\textwidth]{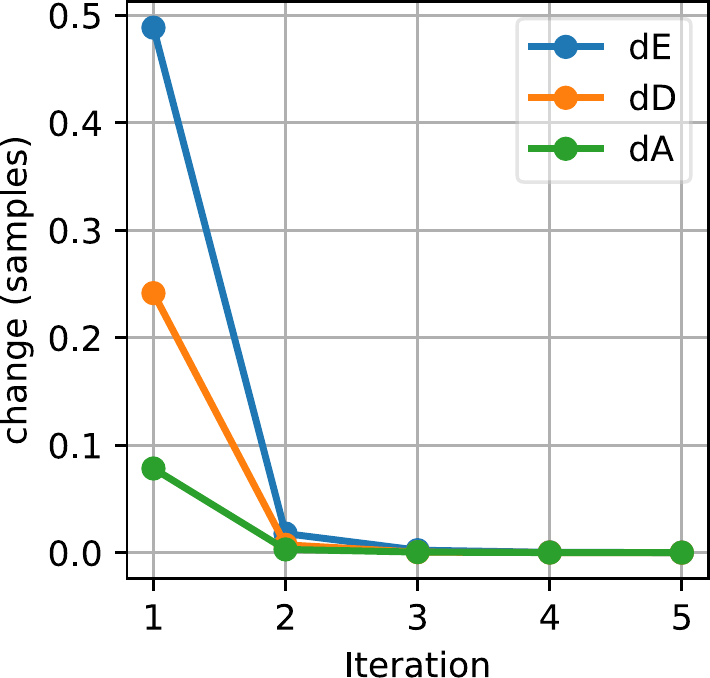}%
\caption{\label{fig:numerical-results-iters}
\begRevTwo%
Convergence of MOB and K-PROX on Einstein. Top to bottom, left to right: MDL cost function; execution time; change in the arguments $A$,$D$ and $E$ for MOB; same for K-PROX. Both methods converge quickly, with K-PROX reaching its near-optimum in just one iteration. In terms of execution time, however, just one iteration of K-PROX required $0.6$s (this is much larger than the $0.11$s reported in Figure~\ref{fig:numerical-results-atoms} since the model has to be learned from scratch), whereas all $10$ MOB  iterations take an accumulated time smaller than $0.1$s. Note that the convergence is exact, not asymptotic: both algorithms stop when  the change of all arguments is $0$. \endRevTwo}
\end{figure}

\begRevTwo
In this set of experiments we are interested mainly in three aspects of the proposed methods. First, the convergence of the forward selection method in \mnote{1.3}\mnote{2.4-5}terms of overall cost function, the convergence rate of the dictionary learning algorithms, and the empirical computational complexity of both the selection and learning methods as measured in running time (seconds), both per iteration and accumulative.\footnote{The timing results were obtained on a Lenovo V310-14IKB notebook with an Intel i5-7200U (4 cores) processor, 8GB of RAM, running Lubuntu 18.04 64bits, with executables compiled from C++ code using GCC 7.3.0 with maximum optimization (``-O3''), multicore support (``-fopenmp'') and all SIMD functions enabled.} %
Figure~\ref{fig:numerical-results-atoms} shows that forward selection converges exponentially to a local minimum which is very close in cost function for both MOB and K-PROX. The forward selection mechanics shown in the break-down of the arguments are typical of methods such as MDL, which shows the correctness of the procedure. 
Figure~\ref{fig:numerical-results-iters} shows that both MOB and K-PROX converge quickly both in cost function and the respective optimization variables (we recall that convergence is exact here -- there is no further change in the arguments). This same behavior was observed for all model sizes in Figure~\ref{fig:numerical-results-atoms}. 
In terms of execution time, both Figure~\ref{fig:numerical-results-atoms} and ~\ref{fig:numerical-results-iters} show MOB as the fastest of the two methods in terms of computing speed. It should be noted however that our implementation for MOB is reasonably optimized, whereas that of K-PROX is not. By comparing the average time for learning a fixed order model in figures~\ref{fig:numerical-results-atoms} and \ref{fig:numerical-results-iters} it should be clear that the warm-restarts strategy used is crucial for efficiently searching through the different model sizes.
\endRevTwo

\section{Concluding remarks}
\label{sec:conclusion}

\begRevTwo 

In this paper we have presented two novel and efficient Binary Factorization Methods based in dictionary learning techniques through the combination of three novel algorithms, BMP for learning coefficients, and MOB and K-PROX for updating the dictionaries. We have provided theoretical guarantees on their convergence to local minima, and a simplified but rigorous analysis of their computational complexity.
\mnote{1.3}\mnote{2.4-5}Through experimentation on two very different datasets, we have demonstrated that our methods produce interpretable results in both cases, requiring very few iterations to converge, and with an overall computational complexity which is very competitive (even though we did not employ any speed-up strategies such as online or mini-batch learning), requiring as little as $0.1$s to learn a complete dictionary on a dataset of $10000$ $289$-dimensional samples from scratch. We have also shown the scalability of our model in terms of growing model size, allowing us to search over a large family (hundreds) of candidate models (dictionaries) in as little as $4$ seconds on a modest notebook. 
In a follow up of this work, which is currently underway, we will present on-line implementations of the MOB and K-PROX algorithms developed here, with the objective of employing them on huge genomic datasets. Other possible lines of work include finding conditions under which our methods (or some of them) can be guaranteed to recover a given underlying model.\mnote{2.2-3}
\endRevTwo

\bibliographystyle{IEEEtran}
\bibliography{IEEEabrv,bmf}
\balance

\end{document}